\documentclass[8pt,a4paper,twocolumn]{article}

\usepackage{graphicx} 
\usepackage{siunitx}

\usepackage{helvet}

\usepackage[utf8]{inputenc}

\usepackage[font=footnotesize, skip = 1pt, labelfont=bf]{caption}
\usepackage[font=footnotesize]{subcaption}

\usepackage{url}

\usepackage{dcolumn}           
\newcolumntype{.}   {D{.}{.}{-1}} 
\newcolumntype{d}[1]{D{.}{.}{#1}} 
\newcolumntype{e}   {D{E}{E}{-1}} 
\newcolumntype{E}[1]{D{E}{E}{#1}} 

\usepackage{amsmath}
\DeclareMathSizes{7}{7}{3}{3} 
\usepackage{pifont}
\usepackage{amsfonts}
\usepackage{amssymb}
\usepackage{amsthm}

\usepackage{float}

\usepackage{sectsty}
\newcommand{\myFontSize}{\fontsize{9}{0}\selectfont}
\sectionfont{\myFontSize}       
\subsectionfont{\myFontSize} 

\usepackage{titlesec}
\usepackage{booktabs}
\usepackage{csquotes}

\titlespacing*{\section}{0pt}{10pt}{0pt}
\titlespacing*{\subsection}{0pt}{10pt}{0pt}

\usepackage{secdot}
\usepackage{epstopdf}
\sectiondot{subsection}
\sectionpunct{section}{. }    
\sectionpunct{subsection}{. } 

\title{\bfseries Detection and Segmentation of Pancreas using Morphological Snakes and Deep Convolutional Neural Networks}
\date{January 2021}
\author{Agapi Davradou \\ adavradou@gmail.com \\ \\ Instituto Superior Técnico, Lisboa, Portugal}

\begin{document}

%
\twocolumn[
\begin{@twocolumnfalse}
\maketitle


\begin{abstract}

Pancreatic cancer is one of the deadliest types of cancer, with 25\% of the diagnosed patients surviving for only one year and 6\% of them for five. Computed tomography (CT) screening trials have played a key role in improving early detection of pancreatic cancer, which has shown significant improvement in patient survival rates. 
However, advanced analysis of such images often requires manual segmentation of the pancreas, which is a time-consuming task. Moreover, pancreas presents high variability in shape, while occupying only a very small area of the entire abdominal CT scans, which increases the complexity of the problem. The rapid development of deep learning can contribute to offering robust algorithms that provide inexpensive, accurate, and user-independent segmentation results that can guide the domain experts.
This dissertation addresses this task by investigating a two-step approach for pancreas segmentation, by assisting the task with a prior rough localization or detection of pancreas. This rough localization of the pancreas is provided by an estimated probability map and the detection task is achieved by using the YOLOv4 deep learning algorithm. The segmentation task is tackled by a modified U-Net model applied on cropped data, as well as by using a morphological active contours algorithm. For comparison, the U-Net model was also applied on the full CT images, which provide a coarse pancreas segmentation to serve as reference.
Experimental results of the detection network on the National Institutes of Health (NIH) dataset and the pancreas tumour task dataset within the Medical Segmentation Decathlon show 50.67\% mean Average Precision. The best segmentation network achieved good segmentation results on the NIH dataset, reaching 67.67\% Dice score. 
\\
\noindent{{\bf Keywords:}} Pancreatic cancer, morphological snakes, deep learning, segmentation, convolutional neural networks \\

\end{abstract}

\end{@twocolumnfalse}
]
	\section{Introduction}
\label{sec:intro}

Cancer is the second leading cause of death worldwide, accounting for an estimated 9.6 million deaths in 2018 \cite{whofactsheet}. Globally, about 1 in 6 deaths is caused by cancer and pancreatic cancer is the seventh highest cause of death from it \cite{whofactsheet}. In most cases, the symptoms are not visible until the disease has reached an advanced stage. Due to its very poor prognosis, after diagnosis, 25\% of people survive only one year and the 5-year survival rate is only 6\% \cite{acs2010cancer}. 

This highlights the need for improved screening modalities and early detection. In this context, 
radiomics, the process of applying machine learning algorithms to extract meaningful information and automate image analysis processes from computed tomography (CT) scans or magnetic resonance imaging (MRI), has become a very active research area. This allows the noninvasive characterization of lesions and the assessment of their progression and possible response to therapies. However, the application of such techniques, besides requiring high quality and reproducible data, often require manual segmentation of the structures of interest which is a time-consuming task. Moreover, manual segmentation is also user-dependent. Consequently, it benefits from automatic segmentation techniques. 

Automatic pancreas segmentation still remains a very challenging task, as the target often occupies a very small fraction (e.g., $<$ 0.5\% of an abdominal CT) of the entire volume, has poorly-defined boundaries with respect to other tissues  and suffers from high variability in shape and size.

The goal of this dissertation is to develop a model to automate pancreas segmentation in CT scans, in order to help the medical community to produce faster and more consistent results.


\par

The next section describes related work, while Section \ref{sec: Methodology} presents the proposed methods and implementation details. Section \ref{sec:resul} details
the obtained results. Finally, Section \ref{sec:concl} summarizes the conclusions and presents directions for future work.


\section{Related Work}
\label{sec: backg}

\subsection{Deep Learning for Pancreas Segmentation}
\label{subsec:procedimiento-experimental}

Early work on pancreas segmentation from abdominal CT used statistical shape models or multi-atlas techniques. In these approaches, the Dice similarity coefficient (DSC) or Dice score in the public National Institutes of Health (NIH) would not exceed 75\%. Therefore, convolutional neural networks (CNNs) have rapidly become the mainstream methodology for medical image segmentation. Despite their good representational power, it was observed that such deep segmentation networks are easily disrupted by the varying contents in the background regions, when detecting small organs such as the pancreas, and as a result produce less satisfying results. Taking that into consideration, a coarse-to-fine approach is commonly adopted. These cascaded frameworks extract regions of interest (RoIs) and make dense predictions on that particular RoIs. More specifically, the state-of-the art methods primarily fall into two categories. 

The first category is based on segmentation networks originally designed for 2D images, such as the fully convolutional networks and the U-Net \cite{ronneberger2015u}. The U-Net architecture utilizes up-convolutions to make use of low-level convolutional feature maps by projecting them back to the original image size, which delineates object boundaries with details. Many frameworks have used a variant of the U-Net architecture to segment the pancreas \cite{lu2019pancreatic}, \cite{man2019deep} and others have made use of fully convolutional network (FCN) and U-Net in order to build more complex models. For example, Zhou et al.\cite{zhou2017fixed} finds the rough pancreas region and then a trained FCN-based fixed-point model refines the pancreas region iteratively. Roth et al. \cite{roth2018spatial} first segment pancreas regions by holistically-nested networks and then refines them by the boundary maps obtained by robust spatial aggregation using random forests. In addition, the TernaryNet proposed by Heinrich et al. \cite{heinrich2018ternarynet} applies sparse convolutions to the CT pancreas segmentation problem, which reduce the computational complexity by requiring a lower number of non-zero parameters. Combinations of convolutional neural networks with recurrent neural networks \cite{yu2018recurrent}, \cite{yang2019pancreas}, \cite{cai2018pancreas} have also been applied, as well as long short-term memory (LSTM) networks \cite{li2019pancreas}. Recently, Zheng et al. \cite{zheng2020deep} proposed a model, which can involve uncertainties in the process of segmentation iteratively, by utilizing the shadowed sets theory. In some cases, in order to incorporate spatial 3D contextual information through the 2D segmentation, slices along different views (axial, sagittal, and coronal) are used, by fusing the results of all 2D networks, e.g. through majority voting. 

In the second category, the methods are based on 3D convolutional layers and therefore operations such as 2D convolution, 2D max-pooling, and 2D up-convolution are replaced by their 3D counterparts. Such networks are the 3D U-Net \cite{cciccek20163d} (which is a 3D extension of the U-Net), the Dense V-Net \cite{gibson2018automatic}, ResDSN \cite{zhu20183d}, 3DFCN \cite{roth2017hierarchical}, and more \cite{zhao2019fully}. In addition, OBELISK-Net proposed by Heinrich et al. \cite{heinrich2019obelisk} applies sparse convolutions to 3D U-Net, Oktay et al.  \cite{oktay2018attention} applies attention gates (AG) and Khosravan et al. \cite{khosravan2019pan} utilizes Projective Adversarial Network to perform pancreas segmentation. Finally, Zhu et al. \cite{zhu2019v} applies neural architecture search, to automatically find optimal network architectures between the 2D, 3D and pseudo3D convolutions.

\subsection{Active Contours and Morphological Snakes}
\label{subsec:materiales-sustancias}

Active contours, also known as "snakes", were first introduced in 1988 by Kass et al. \cite{kass1988snakes}. These are energy-minimizing methods, which are used extensively in medical image processing, as a segmentation technique.
The idea is to initialize a position for the contour, and then define image forces that act on the contour, making it change its position and adapt to the image's features. For instance, Kass et al. \cite{kass1988snakes} define three energies: the internal energy ($E_{int}$), the image energy ($E_{image}$) and the constraint energy ($E_{con}$),  which represent the internal energy of the snake due to bending, the image's energy (which takes into consideration, for example, edges and lines) and the energy of external constraint forces (which takes into consideration constraints created by the user), respectively. In this work, the authors propose different expressions for each energy, which are beyond the scope of this work and will not be presented. 
After defining all the energies, Kass et al. \cite{kass1988snakes} present their active contour algorithm as the solution of the following energy minimization problem: 

\begin{equation}
	\begin{aligned}
E_{snake}^* &:=  \int_0^1 E_{snake}\big(v(s)\big) \hspace{0.1cm} ds \\ 
&:= \int_0^1 E_{int}\big(v(s)\big) \\
&+ E_{image}\big(v(s)\big) + E_{con}\big(v(s)\big) \hspace{0.1cm} ds,
	\end{aligned}
\end{equation}
where $\displaystyle \int_C f(r) \hspace{0.1cm} ds := \int_a^b f\big(r(t)\big) \hspace{0.1cm} \big|r'(t)\big| \hspace{0.1cm} dt $ is the line integral of $f$ along a piecewise smooth curve $C$, $r:[a,b] \rightarrow C$ an arbitrary bijective parametrization of the curve $C$ such that $r(a)$ and $r(b)$ are the endpoints of $C$, with $a<b$. 

Since the minimization of energy leads to dynamic behavior in the segmentation and because of the way the contours slither while minimizing energy, Kass et al. \cite{kass1988snakes} called them snakes. It should also be noted that this minimization problem is generally not convex, which means that different snake initializations may lead to different segmentations.
In order to overcome the problems of bad initialization and local minima, many variants of this method have been proposed, such as using a balloon force  to encourage the contour expansion \cite{cohen1991active} or incorporating gradient flows \cite{kichenassamy1995gradient}. Geometric models for active contours were also introduced, by Caselles et al. \cite{caselles1993geometric}, Yezzi et al. \cite{yezzi1997geometric}, and more \cite{sundaramoorthi2007sobolev}, \cite{malladi1995shape}, as well as a geodesic active contour (GAC) model by Caselles et al. \cite{caselles1995geodesic}. \cite{cremers2001diffusion} et al. incorporated statistical shape knowledge in a single energy functional, by modifying the Mumford-Shah functional \cite{mumford1989optimal} and its cartoon limit and Chan and Vese \cite{chan2001active} introduced a region-based method which is not using an edge-function, but it minimizes an energy, which can be seen as a particular case of the minimal partition problem. Additional region-based methods were proposed, such as those by Li et al. \cite{li2008minimization}, Zhu and Yuille \cite{zhu1996region}, and Tsai et al. \cite{tsai2001curve}.

Inspired by the active contour evolution, a new framework for image segmentation was proposed in \cite{alvarez2010morphological}, which the authors called Morphological Snakes. Instead of computing the snake that minimizes $E_{snake}^*$, Alvarez et al. \cite{alvarez2010morphological} proposed a new approach, which focuses on finding the solution of certain set of partial differential equations (PDEs). 
This approach also yields a snake-like curve and because it approximates the numerical solution of the standard PDE snake model by the successive application of a set of morphological operators (such as dilation or erosion) defined on a binary level-set, it results in a much simpler, faster and stable, curve evolution.

The same authors \cite{marquez2013morphological} have introduced morphological versions of two of the most popular curve evolution algorithms: morphological active contours without edges (MorphACWE) \cite{chan2001active} and morphological geodesic active contours (MorphGAC) \cite{caselles1995geodesic}. 

\par MorphACWE works well when pixel values of the inside and the outside regions of the object to segment have different average. It does not require that the contours of the object are well defined, and it can work over the original image without any preprocessing.
\par MorphGAC is suitable for images with visible contours, even when these contours are noisy, cluttered, or partially unclear. It requires, however, that the image is preprocessed to highlight the contours and the quality of the MorphGAC segmentation depends greatly on this preprocessing step. 

Considering the application of this work, the MorphGAC algorithm was adopted 


	\section{Methodology}
\label{sec: Methodology}

\subsection{Datasets} 
\label{datasets_subsection}

This work relied on two datasets, the pancreas tumour task dataset within the Medical Segmentation Decathlon \cite{simpson2019large} and the NIH (National Institutes of Health) dataset \cite{nih2016data}, \cite{roth2015deeporgan}, \cite{clark2013cancer}.

The Decathlon dataset is a recent multi-institutional effort to generate a large, open-source collection of annotated medical image datasets of various clinically relevant anatomies. The pancreas dataset was collected by the Memorial Sloan Kettering Cancer Center (New York, NY, USA) and contains 420 3D CT scans, 282 for training and 139 for testing. It consists of patients undergoing resection of pancreatic masses (intraductal mucinous neoplasms, pancreatic neuroendocrine tumours, or pancreatic ductal adenocarcinoma). The CT scans have slice thickness of 2.5 mm and a resolution of 512$\times$512 pixels. An expert abdominal radiologist performed manual slice-by-slice segmentation of the pancreatic parenchyma and pancreatic mass (cyst or tumour), using the Scout application \cite{dawant2007semi}.

The NIH dataset contains 80 abdominal contrast-enhanced 3D CT scans from 53 male and 27 female subjects. It consists of healthy patients, since 17 of them are healthy kidney donors scanned prior to nephrectomy and the remaining 65 subjects were selected by a radiologist from patients who neither had major abdominal pathologies nor pancreatic cancer lesions. The CT scans have slice thickness between 1.5 and 2.5 mm and a resolution of 512$\times$512 pixels, with varying pixel sizes. The pancreas was manually segmented in each slice by a medical student and the segmentations were verified/modified by an experienced radiologist.

Due to computational power limit, only a certain amount of data was used. More specifically, the detection model was trained and evaluated using both datasets and the segmentation model was trained and evaluated using only the NIH dataset. More details are given in Sections \ref{subsection_my_yolov4} and \ref{subsection_my_unet}, respectively. In addition, only the information from the transversal plane was taken into consideration. In order to evaluate the variability in the data, a probability map was recreated to define the most likely position of the pancreas. Moreover, the average Hounsfield range and percentage of the volume occupied by the pancreas was also evaluated. The results are presented in Section \ref{section_data_exploration}. 

%

\subsection{Detection Network Architecture}

\subsubsection{Pre-processing}

Pre-processing is a crucial step in machine learning in order to improve the performance of the models. For the detection task, the intensity of all data was clipped between -200 to +300 HU, in order to capture the pancreas intensity range and intensify its boundaries. More specifically, this range was carefully chosen considering the data exploration results that will be later presented in Section \ref{section_data_exploration}.
The -125 to +225 intensity range was also tested for clipping, as well as using 16-bit input images, in order to provide the algorithm with more information. However, both proved to be inefficient and were not implemented.

\subsubsection{YOLOv4}
\label{subsection_my_yolov4}

In this work, the publicly available YOLOv4 \cite{bochkovskiy2020yolov4} was deployed as a detection network. The detection network was trained with both the Decathlon and the NIH datasets. These datasets were shuffled and split into training and validation sets. More specifically, 66 scans from NIH and 34 scans from Decathlon (100 in total) were used for training with 20\% validation. The hold-out method was used to evaluate the performance of the model, by using 10\% of the scans (5 of each dataset) for testing.

Relying on studies regarding the effectiveness of transfer learning in deep networks \cite{pan2009survey}, \cite{shin2016deep}, weights pre-trained on the ImageNet dataset \cite{krizhevsky2012imagenet} were used. Afterwards, YOLOv4 was fine-tuned and re-trained with the pancreas images. 
The model was trained for 20000 iterations with a batch size of 64 and a learning rate of 0.0013.

\subsubsection{Evaluation Metrics}
\label{subsection_yolo_evaluation_metrics}

The performance of all models was evaluated through the hold-out methodology, using the corresponding test datasets. For the quality assessment of the predictions, various metrics were used, i.e., precision, recall, IoU, and mean average precision (mAP) for the detection task. All of the metrics are briefly introduced in the following paragraphs.

Precision, also known as positive predictive value (PPV), measures how accurate the model's predictions are, i.e., the percentage of the predictions that are correct. It is given as the ratio of true positives and the total number of predicted positives, as seen in the following expression: 

\begin{equation}
PPV = \frac{TP}{TP + FP},
\end{equation}
where TP denotes the true positives (predicted correctly as positive) and FP the false positives (predicted incorrectly as positive).

Similarly, recall, also known as sensitivity or true positive rate (TPR), measures how well all the positives are predicted. It is given as the ratio of the true positives and the total of ground truth positives, as seen in the following equation: 

\begin{equation}
TPR = \frac{TP}{TP + FN},
\end{equation}
where FN are the false negatives (incorrectly predicted as negatives).

\begin{figure}[H]
\centering
\includegraphics[width=0.32\textwidth]{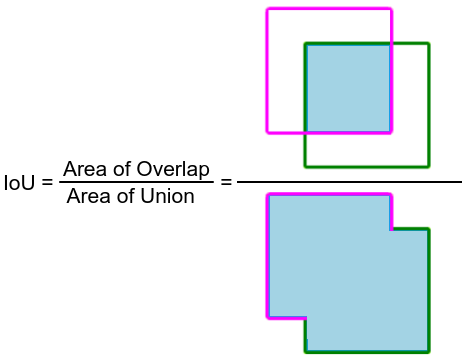}
\caption{Calculation of the IoU metric. With purple color the predicted bounding box is depicted and with green the ground truth.}
\label{fig_IoU_graph}
\end{figure}

The IoU metric, also known as Jaccard index, is a popular similarity measure for object detection problems using the predicted and ground-truth bounding boxes. Evidently, as exemplified in Figure \ref{fig_IoU_graph}, a bigger overlap between the two bounding boxes results in higher IoU score and therefore, detection accuracy.

The average precision (AP) is a default evaluation metric in the PascalVOC competition \cite[p.~313]{everingham2010pascal}, which derives from the area under the precision/recall curve. The precision/recall curve is
computed from a model's ranked output and the predictions are ranked with respect to the confidence score of each bounding box. In this work, the detection model is set to keep only predictions with a confidence score higher than 25\%. The AP provides an indication of the shape of the precision/recall curve, and is defined as the interpolated average precision at a set of eleven equally spaced recall levels [0, 0.1,..., 1], expressed as:

\begin{equation}
AP = \frac{1}{11} \sum_{r \in \{0,0.1,...,1\}}p_{interp}(r).
\end{equation}

At each recall level $r$, the interpolated precision $p_{interp}$ is calculated by taking the maximum precision measured for that $r$, given by the following formula:

\begin{equation}
p_{interp}(r) = \underset{\tilde{r}:\tilde{r}\geq r}{max} \ p(\tilde{r}).
\end{equation}

Since mAP is calculated by taking the average of the AP calculated for all the classes, they will be used interchangeably in the current context.


\subsection{Segmentation Network Architecture}

\subsubsection{Pre-processing}

Intensity clipping was not adopted for the segmentation task, since it did not show any improvement in the performance of the models. For the segmentation network, curvature driven image denoising is applied on each slice, in order to make the pixel distribution more uniform. Finally, contrast limited adaptive histogram equalization (CLAHE) and normalization were also investigated for the segmentation task, but they also proved to be ineffective, in terms of Dice score.

\subsubsection{U-Net model}
\label{subsection_my_unet}

\begin{figure*}[h]
\centering
\includegraphics[width=1.00\textwidth]{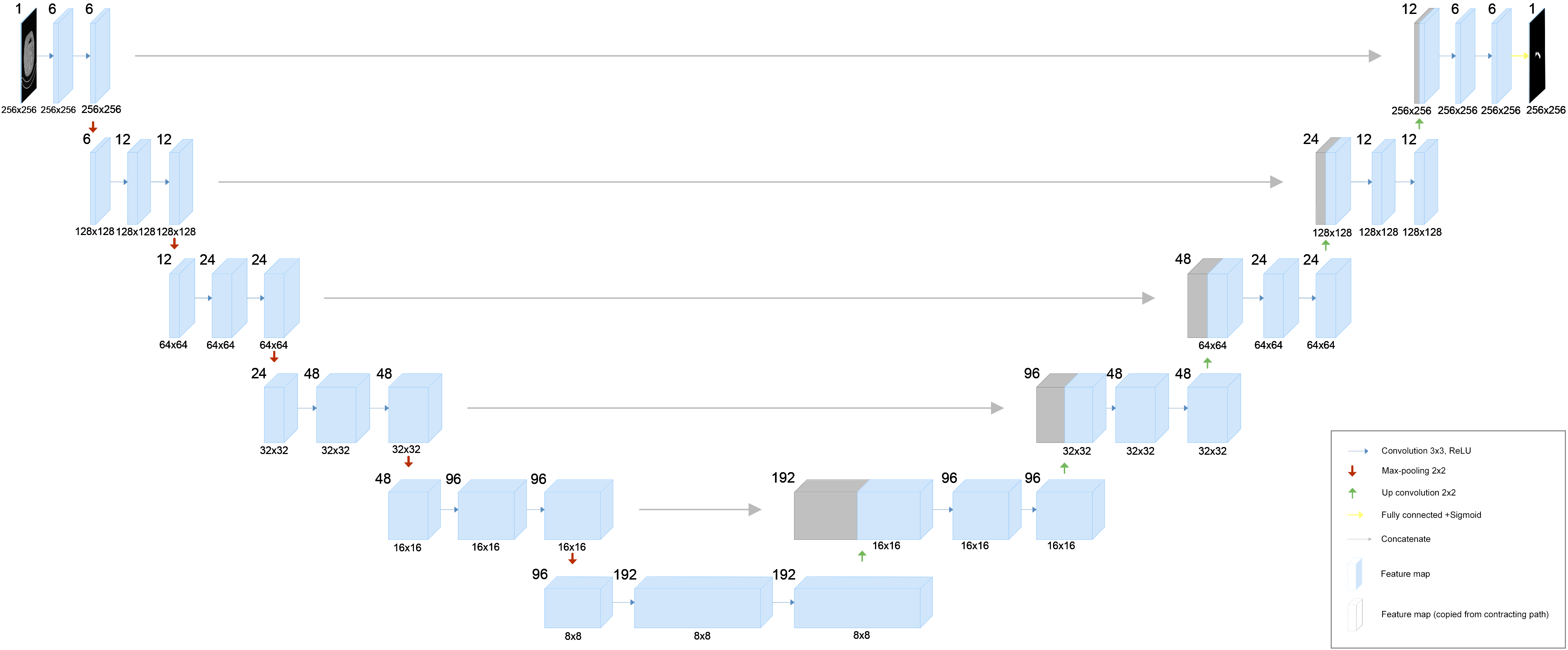}
\caption{The 5-level U-Net segmentation network for an input image of size 256x256.}
\label{fig_my_U-Net}
\end{figure*}

An adaptation of the original U-Net architecture was chosen as a segmentation model. More specifically, following the original implementation, 3$\times$3 convolutional kernels are used in the contracting path with a stride of 1, each followed by a 2$\times$2 max-pooling operation with stride 2, and 2$\times$2 up-convolutional kernels with a stride of 2 are implemented in the expansive path. Each convolution layer, except for the last one, is followed by a ReLU, and dropout of 0.5 is selected. However, the number of levels in both the descending and ascending paths of the network was increased from four blocks to five blocks and the number of filters per layer was reduced compared to the original implementation, as visualized in Figure \ref{fig_my_U-Net}. In addition, residual connections were included in each convolutional block, padding was added and the sigmoid function was used as an activation function for the last layer. Due to memory limitations, all scans were resized from 512$\times$512 to 256$\times$256.

As mentioned in Section \ref{datasets_subsection}, the U-Net model was trained on 50 volumes from the NIH dataset, 45 of them were used for training and 5 for validation. 

Since the pancreas occupies only a very small region of a CT-Scan, the work \cite{milletari2016v} is followed, and a DSC-loss layer is used to prevent the model from being heavily biased towards the background class. In more detail, the DSC between two voxel sets, $A$ and $B$, can be expressed as

\begin{equation}
DSC(A,B) = \frac{2x|A \cap B|}{|A|+|B|},
\end{equation}
and this is slightly modified into a loss function between the ground-truth mask Y and the predicted mask $\hat{Y}$, in the following way: 

\begin{equation}
L(\hat{Y},Y) = 1-\frac{2x \sum_{i}\hat{y}_iy_i + \epsilon}{\sum_{i}\hat{y}_i + \sum_{i}y_i + \epsilon},
\end{equation}
where $\epsilon$ is an added term to avoid underflow. 


The model is trained with batches of 128 instances and optimized using Adam with a learning rate of 0.0001. Early-stopping is also used, in order to avoid overfitting of the network, and the model with the lowest validation loss is chosen.

Data augmentation is also implemented, by applying elastic deformation of images as described in \cite{simard2003best}, as well as image shifting, rotation, zooming and flipping, in order to expand the training dataset and make the network more robust to such variations.

\subsubsection{Cropped U-Net model}
\label{cropped_vanilla_u-net_section}

In order to reduce the irrelevant information in a CT-scan and train the U-Net on scans with more useful information, the scans were cropped to a smaller size. The rest of the training procedure described in Section \ref{subsection_my_unet} was kept intact.

\begin{figure}[H]
\centering
\includegraphics[width=0.28\textwidth]{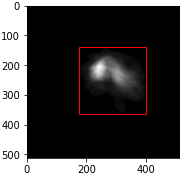}
\caption{Default cropping position of a 512$\times$512 CT-scan into a 224$\times$224 scan.}
\label{fig_Default_Cropping_Position}
\end{figure}

The image size of the cropped scans was carefully decided, as on the one hand it should be as small as possible, but also contain the whole pancreas. Taking into account the information retrieved from Figure \ref{fig_pancreas_probability_map}, which shows the probability of pancreas existing in a 512$\times$512 CT slice, as well as the maximum size of pancreas in both datasets, a default cropping position was decided. The default cropping position has its centroid at position $(x = 287, y = 250)$ and image size of 224$\times$224, as visualized in Figure \ref{fig_Default_Cropping_Position}.

\subsubsection{YOLO + MorphGAC model}
\label{yolo+morphgac_methodology}

In order to overcome the problem of initialization, the MorphGAC algorithm was combined with the YOLO detection model. The MorphGAC segmentation was applied on cropped bounding box predictions using their centroid as an initialization point. The segmentation results are then repositioned back to 512$\times$512 images, in order to form the final 3D segmented pancreas.


\subsubsection{YOLO + U-Net model}
An additional two-step approach with cropped CT-scans is proposed, which combines the YOLO detection model to crop the images containing the pancreas, as well as the default cropping position mentioned in Section \ref{cropped_vanilla_u-net_section}, to crop the images without pancreas. The U-Net architecture visualized in Figure \ref{fig_my_U-Net} and the training procedure described in \ref{subsection_my_unet} were used.

\subsubsection{Post-processing}

\begin{figure}[H]
\centering
\includegraphics[width=0.5\textwidth]{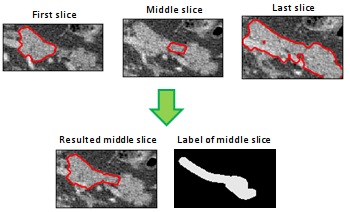}
\caption{Example of the application of the optimization step on a slice. The optimization is applied only on the middle slice of the three used.}
\label{fig_Optimization_example}
\end{figure}

A post-processing step is adopted for all predicted segmentations, aiming for optimization and a smoother result, by leveraging spatial information. Since pancreas has a uniform and undivided shape, it is unlikely that a pixel in a 2D slice has a different value from both the previous and the next slice. Taking that into consideration, all slices in a predicted segmentation are compared in groups of three and the values of the middle ones are changed, when different from the other two. In Figure \ref{fig_Optimization_example}, the segmentation improvement on the middle slice is depicted, after post-processing it. 

\subsubsection{Evaluation Metrics}
\label{subsection_u-net_evaluation_metrics}

The performance of all models was evaluated through the hold-out methodology, using the corresponding test datasets.  Since MorphGAC algorithm needs no training, the YOLO+MorphGAC approach was evaluated on the detection's model test dataset. For the quality assessment of the segmentation task, the DSC or Dice score was used. Because of the similarity between the IoU metric and the Dice score, the former was not used for the evaluation of the segmentation models.

\begin{figure}[H]
\centering
\includegraphics[width=0.4\textwidth]{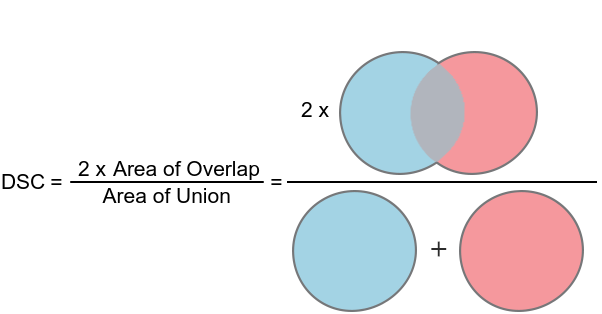}
\caption{Calculation of the DSC metric.}
\label{fig_DSC_graph}
\end{figure}

The Dice score is the most common evaluation metric for the segmentation of medical images and it is calculated as twice the area of overlap divided by the total number of pixels in both images, as illustrated in Figure \ref{fig_DSC_graph}.

Evidently, the DSC is equivalent to the IoU, using the following expression:
 
\begin{equation}
DSC = \frac{2 \mbox{IoU}}{1+\mbox{IoU}}.
\end{equation}

	\section{Results \& discussion}
\label{sec:resul}

\subsection{Data Exploration}
\label{section_data_exploration}

\begin{figure}[h]
\centering
\includegraphics[width=0.45\textwidth]{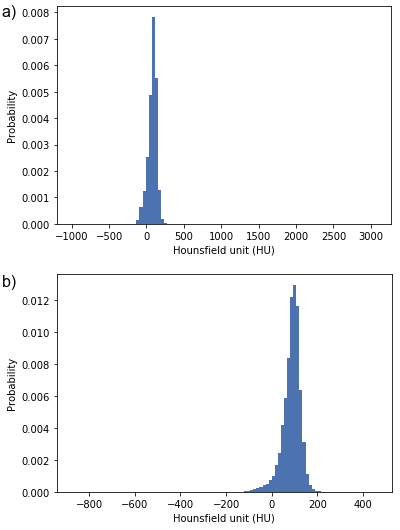}
\caption{The Hounsfield unit values of the pancreas for the Decathlon (a) and NIH (b) dataset, respectively. In Decathlon dataset the intensity range from -998.0 HU to +3071.0, while in NIH from -875.0 HU to +458.0 HU.}
\label{fig_pancreas_HU_intensities}
\end{figure}

The average Hounsfield range and percentage of the volume occupied by the pancreas were evaluated. In Figure \ref{fig_pancreas_HU_intensities} the pancreas intensity values for both datasets are illustrated, where Decathlon has a mean +80.63 $\pm$ 57.91 HU and NIH +86.71 $\pm$ 32.18 HU. 


\begin{figure}[h]
\centering
\includegraphics[width=0.5\textwidth]{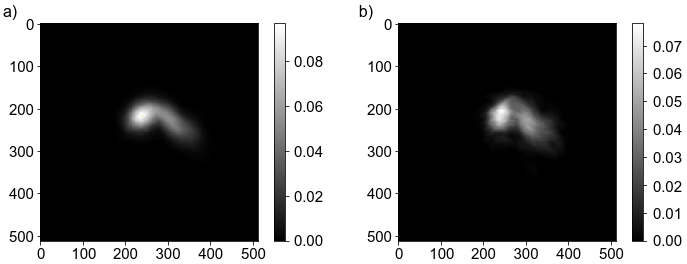}
\caption{Probability of pancreas in an 512x512 image for Decathlon (a) and NIH (b) datasets, respectively.}
\label{fig_pancreas_probability_map}
\end{figure}

In addition, the mean percentage of pancreas in the abdominal ct scans is 0.46\% and 0.49\% for Decathlon and NIH respectively, taking into consideration only values $<$800 HU, in order to exclude the air. Moreover, a probability map was recreated to define the most likely position of the pancreas. In Figure \ref{fig_pancreas_probability_map} is the probability map of pancreas in a 2D slice for both datasets is visualized.  In the Decathlon dataset, the pancreas location ranges from pixel 150 to 434 in the x-axis and from pixel 139 to 348 in the y-axis. Similarly, in NIH dataset, the pancreas location ranges from 167 to 405 in the x-axis and from 143 to 360 in the y-axis.

\subsection{Detection Task}
\label{subsec:section_datection_task}

\begin{figure}[h]
\centering
\includegraphics[width=0.45\textwidth]{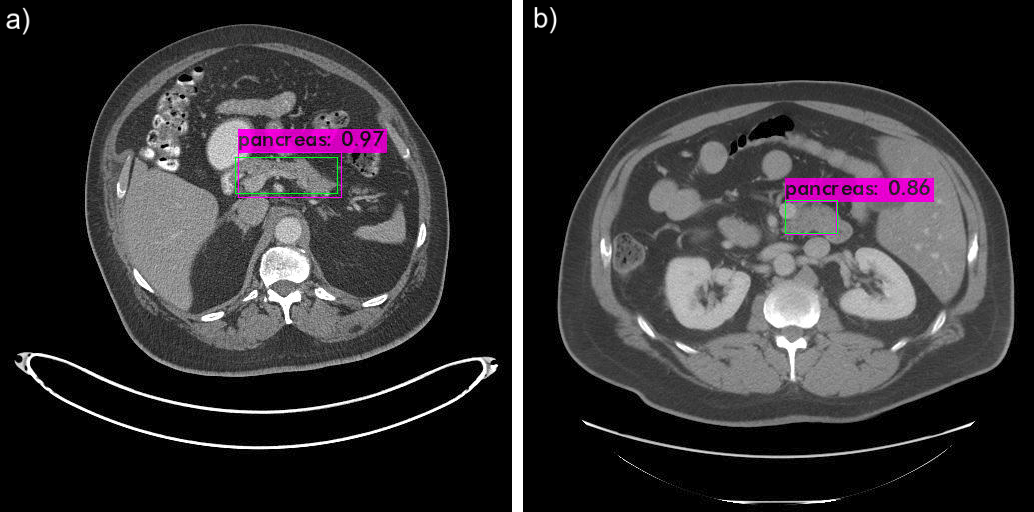}
\caption{Detection of pancreas on a NIH (a) and Decathlon (b) slice, respectively. The green color represents the ground-truth and the purple the YOLO prediction.}
\label{fig_successful_pancreas_detection}
\end{figure}

The detection model was evaluated on the holdout test set containing volumes from both the NIH and the Decathlon datasets and it can predict pancreas successfully in both of them, as visualized in Figure \ref{fig_successful_pancreas_detection}.

The results of the detection model evaluated by the PPV, TPR, IoU, and mAP metrics described in Section \ref{subsection_yolo_evaluation_metrics} are presented in Table \ref{tab_table_detection_results}. The detection model shows a mAP of 50.67\% on the test set, with the IoU being 47.70\%, the precision 0.63, and the recall 0.52. However, the performance of the model varies significantly between the two datasets, with the mAP being 71.43\% for the NIH dataset and 29.92\% for the Decathlon dataset. For that reason, the performance of the model in Table \ref{tab_table_detection_results} is also evaluated separately for each dataset. More specifically, the model shows a IoU of 57.43\% with a precision and recall of 0.63 and 0.52 respectively, on the NIH dataset. On the Decathlon dataset, the model has 37.96\% IoU, the precision is 0.5 and the recall 0.36.

\begin{table}[h]
\centering
\caption{Evaluation results of the detection model.}
\label{tab_table_detection_results}
\scalebox{0.75}{
 \begin{tabular}{ |m{3cm} m{1cm} m{1cm} m{1.5cm} m{1.5cm}| } 
 \hline
   Test dataset & PPV & TPR & IoU & mAP \\ 
 \hline 
   NIH \& Decathlon &  {0.63} & {0.52} & {47.70\%} & {50.67\%} \\ 
   NIH  &  0.75 & 0.68 & 57.43\% & 71.43\% \\ 
   
   Decathlon  & 0.5  & 0.36 & 37.96\% & 29.92\%  \\ 
 \hline  
 \end{tabular}}
\end{table}

\begin{figure}[H]
\centering
\includegraphics[width=0.25\textwidth]{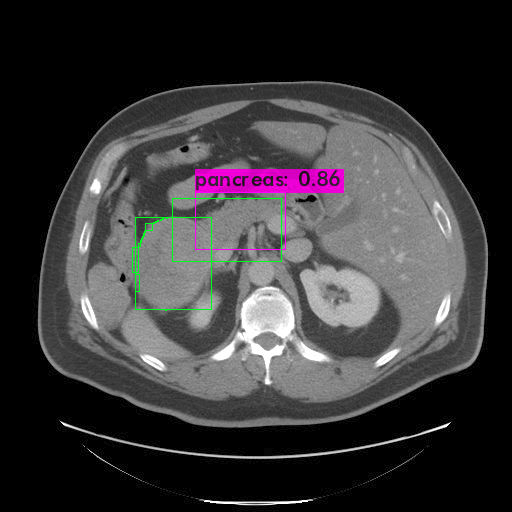}
\caption{Detection of pancreas on a multilabel slice. The green color represents the ground-truth and the purple the YOLO prediction. The two bigger green bounding boxes represent the two classes.}
\label{fig_multilabel_BB_problem}
\end{figure}

The low detection accuracy of the YOLO model on the Decathlon dataset derives from the multi-label nature of some ground-truth files. More specifically, it was later realized the YOLO model was configured to be trained on one class and therefore, since the pancreatic parenchyma and pancreatic mass (cyst or tumour) are annotated separately as different classes in the Decathlon dataset, the second class was ignored.  as shown in Figure \ref{fig_multilabel_BB_problem}. In addition, during the conversion from NIfTI to tiff format (which is required by YOLO), all the individual globs of healthy pancreas surrounding the pancreatic mass, are annotated as separate tiny bounding boxes. However, as visualized on this figure, it is very interesting to notice that the model outputs a fairly good prediction on the class that it was trained on (upper-right green bounding box).

Unfortunately, the problem was not discovered on time, since the YOLO model was trained only on the NIH dataset at first with healthy pancreas. Nevertheless, this could be overcome by considering a single class once the goal of the work was to segment the healthy pancreas. Due to the unavailability of the computer on which YOLO was trained afterwards, this could not be realized in the current dissertation.

\begin{figure}[h]
\centering
\includegraphics[width=0.45\textwidth]{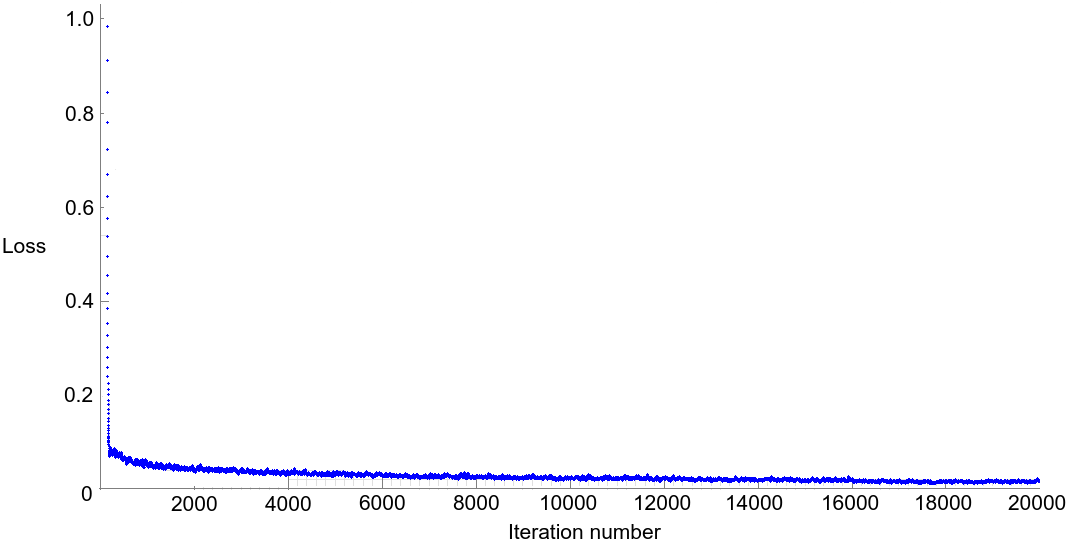}
\caption{The CIoU training loss with a minimum value of 0.18.}
\label{fig_loss_training_yolov4}
\end{figure}

\begin{figure}[h]
\centering
\includegraphics[width=0.45\textwidth]{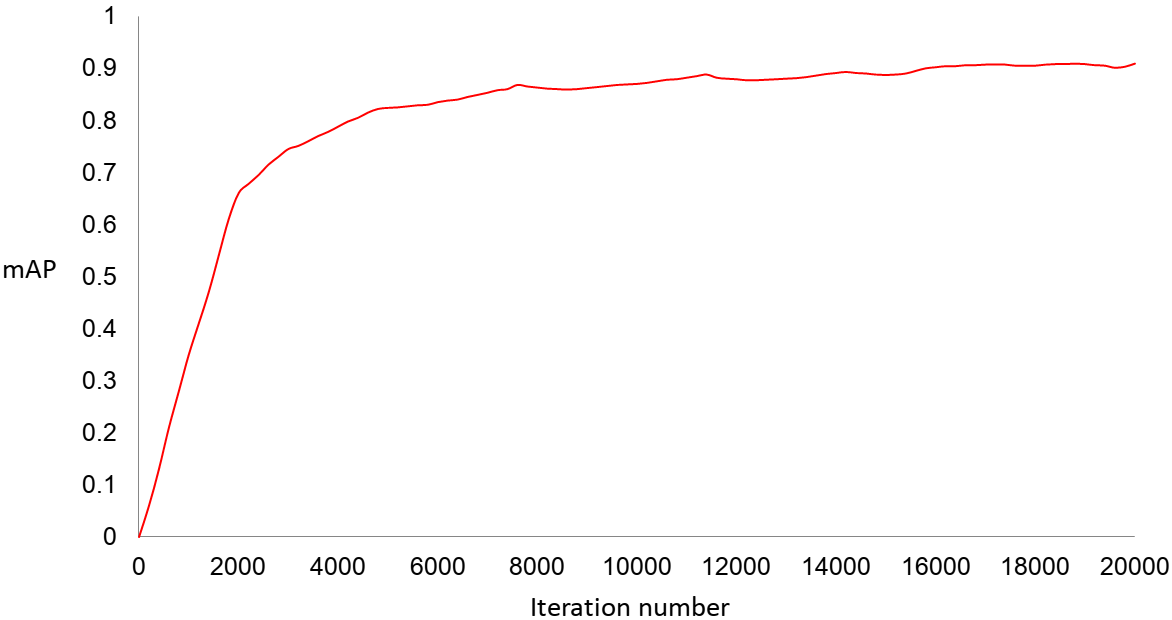}
\caption{The validation mAP during training, reaching a maximum value of 90.9\%.}
\label{fig_mAP_training_yolov4_xls}
\end{figure}

In Figures \ref{fig_loss_training_yolov4} and \ref{fig_mAP_training_yolov4_xls}, the metrics used for the assessment of the detection model during training are visualized.

\subsection{Segmentation Task}
\label{subsec:section_segmentation_task}

\begin{figure}[h]
\centering
\includegraphics[width=0.5\textwidth]{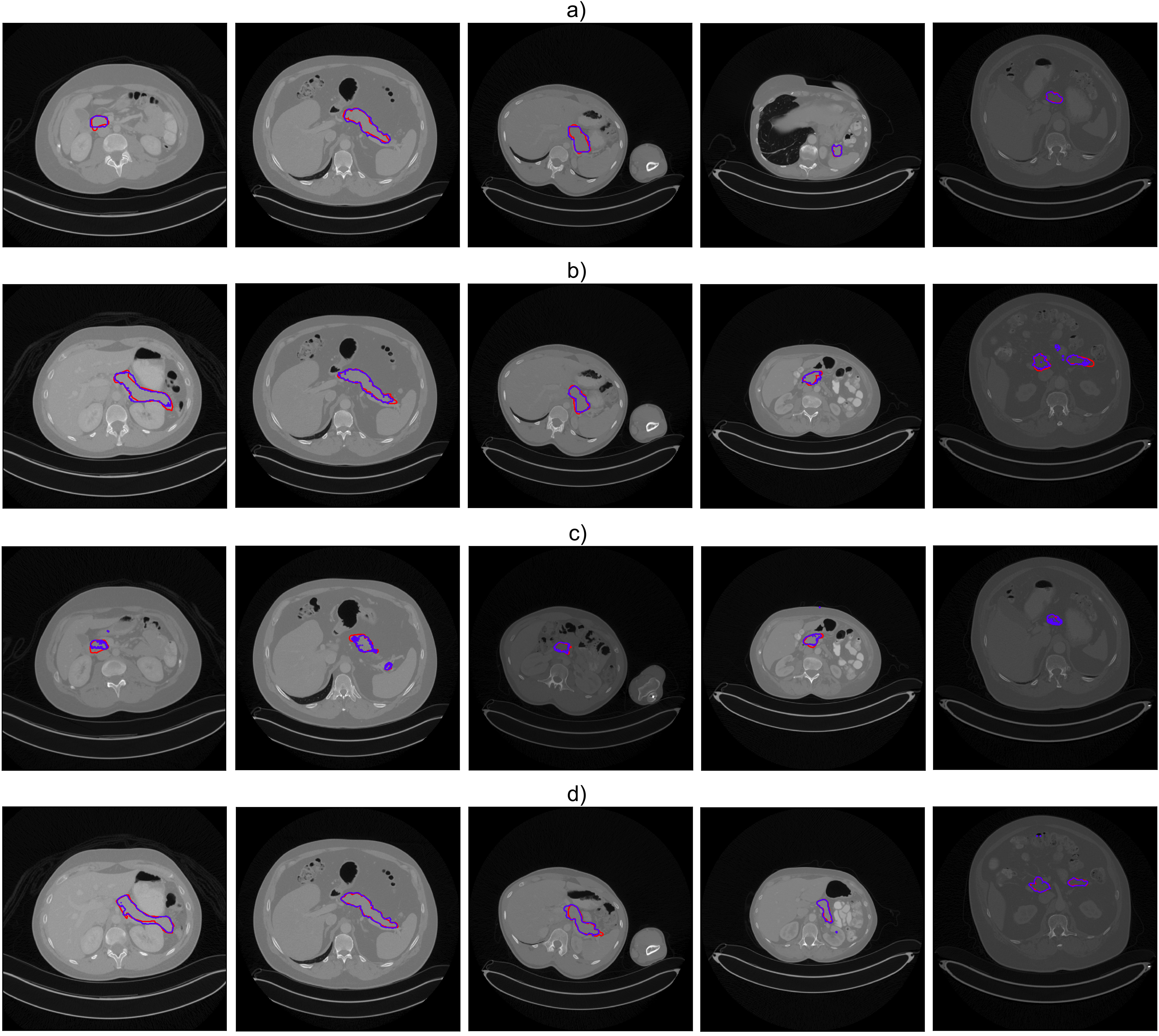}
\caption{Some of the best segmentations on the test set of the YOLO+MorphGAC (a), U-Net (b), cropped U-Net (c) and YOLO+U-Net (d) model, respectively. The blue line indicates the model's predicted segmentation and the red line represents the ground-truth.}
\label{fig_best_segmentation_results}
\end{figure}

The segmentation models were evaluated on the holdout set containing volumes from the NIH dataset and the pancreas can be successfully segmented, as visualized in Figure \ref{fig_best_segmentation_results}.

The segmentation performance of all models on the NIH dataset is presented on Table \ref{tab_overall_segmentation_results}. The YOLO+MorphGAC model has the best performance achieving a dice score of 67.67 $\pm$ 8.62\%, followed by the YOLO+U-Net model with a Dice score of 64.87 $\pm$ 4.79\%. The U-Net and cropped U-Net show a lower performance with a Dice score of 59.91 $\pm$ 13.69\% and 59.08 $\pm$ 12.76\%, respectively. The obtained DSC from the cropped U-Net when compared with the YOLO+U-Net was lower and showed a higher standard deviation. The preliminary results suggest that YOLO was able to improve segmentation when compared to using a probability map. A similar behavior was observed for the U-Net model with improved results when using the YOLO model. Regarding segmentation performance, the highest average Dice score was obtained with the YOLO+MorphGAC model and the lowest standard deviation with the YOLO+U-Net model.

\begin{table}[h]
\centering
\caption{Evaluation results of all proposed segmentation models.}
\label{tab_overall_segmentation_results}
 \begin{tabular}{ |m{3.5cm} m{3.5cm} | } 

 \hline
   Test dataset & Dice score  \\ 
 \hline
 
   {YOLO+MorphGAC} &  {67.67} $\pm$ {8.62 \% }\\ 

   U-Net  &  59.91 $\pm$ 13.69 \% \\ 

   Cropped U-Net  & 59.08 $\pm$ 12.76 \%  \\ 
   
   YOLO+U-Net  & 64.87 $\pm$ 4.79 \%  \\ 
 \hline 
 
 \end{tabular}
\end{table}

\begin{figure}[h]
\centering
\includegraphics[width=0.48\textwidth]{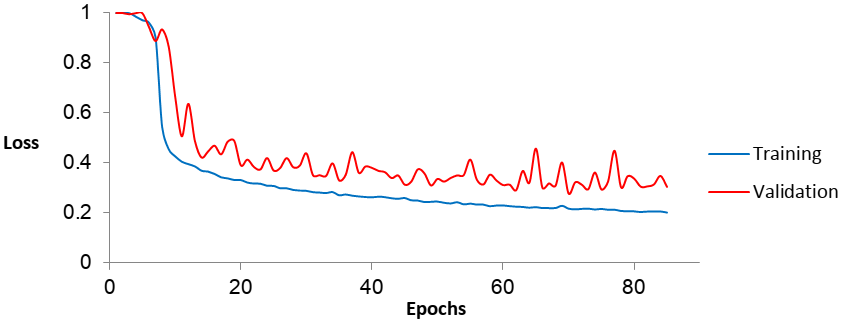}
\caption{The metric of loss for the YOLO+U-Net during training with a minimum value of 0.28.}
\label{fig_YOLO+U-Net_training_graph}
\end{figure}

The loss function used for the assessment of the segmentation model during training is visualized in Figure \ref{fig_YOLO+U-Net_training_graph}.

\begin{figure}[h]
\centering
\includegraphics[width=0.5\textwidth]{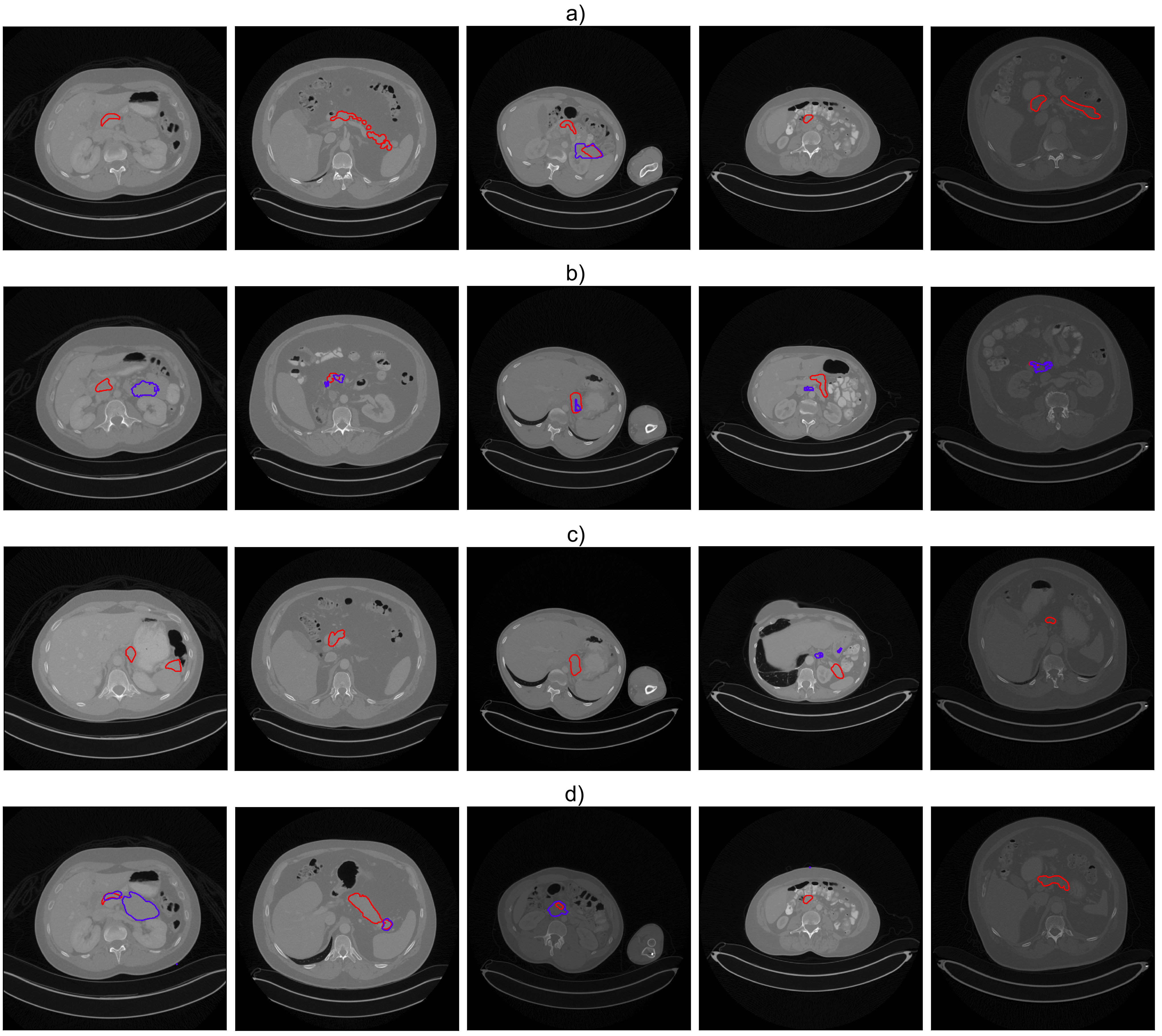}
\caption{Some of the worst segmentations on the test set of the YOLO+MorphGAC (a), U-Net (b), cropped U-Net (c) and YOLO+U-Net (d) model, respectively. The blue line indicates the model's predicted segmentation and the red line represents the ground-truth.}
\label{fig_worst_segmentation_results}
\end{figure}

In Figures \ref{fig_best_segmentation_results} and \ref{fig_worst_segmentation_results} representative slices of the best and worst results of the proposed segmentation models are presented, respectively.

\subsubsection{Comparison with the state-of-the-art}

In Tables \ref{tab_table_state-of-the-art_comparison_dsc} and \ref{tab_table_state-of-the-art_comparison_details} a comparison of the best two proposed models with the state-of-the art deep learning models is presented with respect to the dice coefficient and the training details, respectively. Only models which implement a 2D approach were taken into consideration and trained on the NIH dataset as well. The proposed models display a low performance compared to the other networks. However, when examining the training methodology of all the state-of-the-art methods, it is concluded that cross validation is the adopted validation technique, as well as that they leverage the whole NIH dataset for the training, which in the current work was not possible, due to computational and time limitations. These two reasons could be the main causes for the low segmentation performance of the proposed YOLO+U-Net model and following the state-of-the-art training methods could result in a higher a Dice score.

\begin{table}[h]
\centering
\caption{Comparison of the DSC with the state-of-the-art methods.}
\label{tab_table_state-of-the-art_comparison_dsc}
 \centering
 \scalebox{0.8}{
 \begin{tabular}{ |m{5.5cm} m{3cm}| } 

 \hline
   Method & DSC \\ 
 \hline 
 
   2D U-Net \cite{heinrich2018ternarynet}  & 71.07\% $\pm$ 9.50 \\ 
   
   2D U-Net \cite{man2019deep} & 86.93\% $\pm$ 4.92 \\ 
   
   2D FCN + RNN \cite{cai2018pancreas} & 83.70\% $\pm$ 5.10 \\ 
   
   Holistically Nested 2D FCN \cite{roth2018spatial}  & 81.27\% $\pm$ 6.27\\ 
   
   2D FCN \cite{zhou2017fixed} & 83.18\% $\pm$ 4.81 \\    
   
   YOLO+U-Net (proposed) & 64.87\% $\pm$ 4.79 \\ 
   
   YOLO+MorphGAC (proposed) & 67.67\% $\pm$ 8.62 \\ 
 \hline  
 \end{tabular}}
\end{table}

\begin{table}[h]
\centering
\caption{Comparison of the training details with the state-of-the-art methods.}
\label{tab_table_state-of-the-art_comparison_details}
 \centering
 \scalebox{0.71}{
 \begin{tabular}{ |m{5cm}m{1.3cm} m{1.3cm} m{1.5cm}| } 

 \hline
   Method & Dataset & Train/Test & Validation \\ 
 \hline 
 
   2D U-Net \cite{heinrich2018ternarynet}  & NIH & 63/9 & 5-fold CV  \\ 
   
   2D U-Net \cite{man2019deep}  & NIH & 62/20 & 4-fold CV  \\ 
   
   2D FCN + RNN \cite{cai2018pancreas} & NIH & 62/20 & 4-fold CV  \\ 
   
   Holistically Nested 2D FCN \cite{roth2018spatial}    & NIH & 62/20 & 4-fold CV  \\ 
   
   2D FCN \cite{zhou2017fixed}  & NIH & 62/20 & 4-fold CV  \\    
   
   YOLO+U-Net (proposed) & NIH & 50/5 & holdout  \\ 
   
   YOLO+MorphGAC (proposed) & - & -/5 & -  \\ 
 \hline  
 \end{tabular}}
\end{table}
	\section{Conclusion and Future Work}
\label{sec:concl}

\subsection{Conclusions}

Taking into account the state-of-the-art cascaded segmentation models, the main goal of the current work was to investigate and present simpler two-step approaches for the segmentation of the pancreas in CT. In the end, three models were investigated and compared to the U-Net implementation. The two-step approach was achieved by using a detection network for the pancreas localization prior to the segmentation. This approach, which was combined with both morphological snakes and a U-Net segmentation network, proved to be the most efficient, showing that the organ detection benefits the segmentation task. Another approach was to reduce the background information by cropping the data using a pancreas probability map and then using a U-Net network to segment this smaller area. However, this showed no improvement in segmentation performance, when compared to U-Net segmentation results on whole CT-scans. Due to the Covid-19 situation, many computational limitations existed, which resulted in insufficient training of both the detection and the segmentation network, which is the main reason of the low performance of the proposed models. Since the segmentation performance lies below 70\%, it is concluded that further investigation and model improvement is needed, in order to tackle the challenging nature of the pancreas segmentation problem efficiently.

\subsection{Future Work}

Regarding the proposed models, there are changes that could be investigated to further improve the performance, focusing on YOLO+MorphGAC and YOLO+U-Net architectures. First and foremost, since the detection network plays a vital role on both of them, its improvement would also lead to overall improved performance of the model. Considering that YOLOv4 is one of the most modern and efficient state-of-the-art detection model, experimenting with different ones would not be the main priority. A slight improvement could occur by testing different backbone models for feature extraction, such as ResNet \cite{he2016deep} or EfficientNet \cite{tan2019efficientnet}. More importantly, leveraging the whole Decathlon dataset and training on a larger dataset, could also lead to a better detection performance. Regarding the proposed U-Net architecture, similarly to the detection task, the need to train on a larger dataset is undeniable and adopt the training methodology of the state-of-the-art models. In addition, a YOLO+FCN approach would also be interesting for investigation. Finally, exploring 3D versions of the architectures could also improve the results. The present work focuses on the segmentation of healthy pancreas, nevertheless, once Decathlon dataset also presents tumor masks, tumor segmentation could also be explored.


%
\bibliographystyle{abbrv}

\bibliography{references}

\end{document}